\documentclass[10pt,twocolumn,letterpaper]{article}

\usepackage{cvpr}
\usepackage{times}
\usepackage{epsfig}
\usepackage{graphicx}
\usepackage{amsmath}
\usepackage{amssymb}
\usepackage{bm}
\usepackage{multirow}
\usepackage{booktabs}
\usepackage{authblk}

\usepackage[pagebackref=true,breaklinks=true,colorlinks,bookmarks=false]{hyperref}

\cvprfinalcopy 

\ifcvprfinal\fi

\usepackage{enumitem}
\setitemize[1]{itemsep=-0.2em,topsep=0em}
\usepackage{graphicx}
\usepackage{subfigure}
\usepackage[font=small]{caption}
\usepackage{setspace}
\begin{document}

\title{FOTS: Fast Oriented Text Spotting with a Unified Network\vspace{-1cm}}

\author[1]{\normalsize Xuebo Liu}
\author[1]{Ding Liang}
\author[1]{Shi Yan}
\author[1]{Dagui Chen}
\author[2]{Yu Qiao}
\author[1]{Junjie Yan\vspace{-0.3cm}}
\affil[1]{SenseTime Group Ltd.}
\affil[2]{Shenzhen Institutes of Advanced Technology, Chinese Academy of Sciences}
\affil[ ]{\tt\small {\{liuxuebo,liangding,yanshi,chendagui,yanjunjie\}@sensetime.com}, {\{yu.qiao\}@siat.ac.cn}}

\maketitle
\thispagestyle{empty}

\begin{abstract}
  Incidental scene text spotting is considered one of the most difficult and valuable challenges in the document analysis community. Most existing methods treat text detection and recognition as separate tasks. In this work, we propose a unified end-to-end trainable Fast Oriented Text Spotting (FOTS) network for simultaneous detection and recognition, sharing computation and visual information among the two complementary tasks. Specially, RoIRotate is introduced to share convolutional features between detection and recognition. Benefiting from convolution sharing strategy, our FOTS has little computation overhead compared to baseline text detection network, and the joint training method learns more generic features to make our method perform better than these two-stage methods. Experiments on ICDAR 2015, ICDAR 2017 MLT, and ICDAR 2013 datasets demonstrate that the proposed method outperforms state-of-the-art methods significantly, which further allows us to develop the first real-time oriented text spotting system which surpasses all previous state-of-the-art results by more than 5\% on ICDAR 2015 text spotting task while keeping 22.6 fps.

\end{abstract}

\section{Introduction}

Reading text in natural images has attracted increasing attention in the computer vision community \cite{tian2016ctpn,shi2017seglink,zhou2017east,shi2016crnn,he2016first,he2017casia,liao2017textboxes}, due to its numerous practical applications in document analysis, scene understanding, robot navigation, and image retrieval. Although previous works have made significant progress in both text detection and text recognition, it is still challenging due to the large variance of text patterns and highly complicated background.

\begin{figure}
  \includegraphics[width=0.47\textwidth]{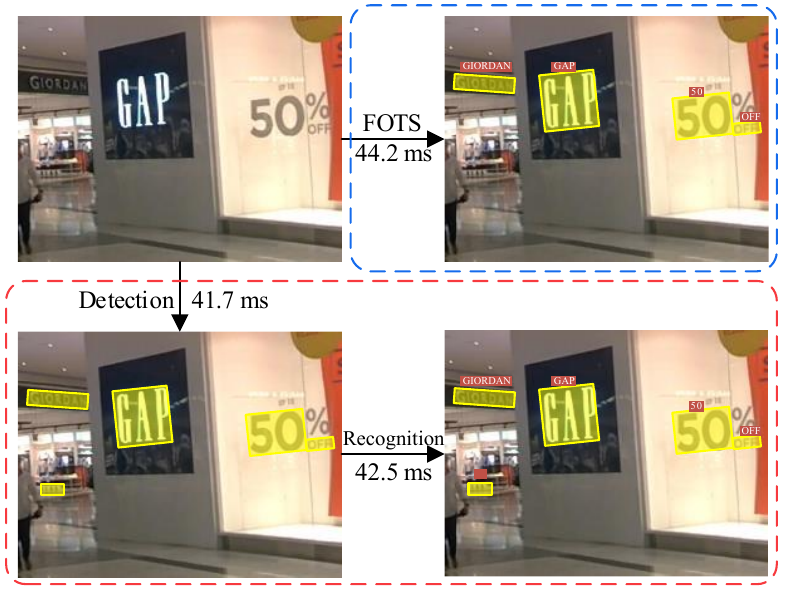}
  \vspace{5pt}
  \caption{Different to previous two-stage methods, FOTS solves oriented text spotting problem straightforward and efficiently. FOTS can detect and recognize text simultaneously with little computation cost compared to a single text detection network (44.2ms vs. 41.7ms) and almost twice as fast as the two-stage method (44.2ms vs. 84.2ms). This is detailed in Sec. \ref{speed}.}
  \label{fig:fastdemo}
\end{figure}

The most common way in scene text reading is to divide it into text detection and text recognition, which are handled as two separate tasks \cite{jaderberg2016ijcv, liao2017textboxes}. Deep learning based approaches become dominate in both parts. In text detection, usually a convolutional neural network is used to extract feature maps from a scene image, and then different decoders are used to decode the regions \cite{tian2016ctpn,shi2017seglink,zhou2017east}. While in text recognition, a network for sequential prediction is conducted on top of text regions, one by one \cite{shi2016crnn, he2016first}. It leads to heavy time cost especially for images with a number of text regions. Another problem is that it ignores the correlation in visual cues shared in detection and recognition. A single detection network cannot be supervised by labels from text recognition, and vice versa.

\begin{figure*}
  \includegraphics[width=\textwidth]{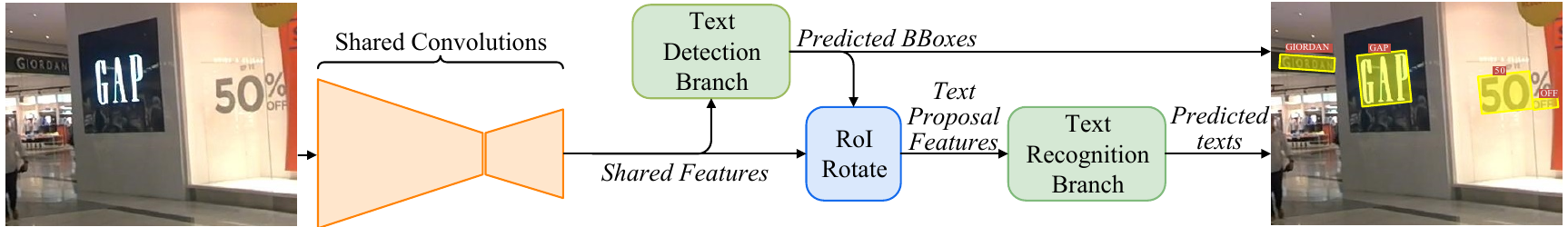}
  \vspace{0pt}
  \caption{Overall architecture. The network predicts both text regions and text labels in a single forward pass.}
  \label{fig:architecture}
\end{figure*}

In this paper, we propose to simultaneously consider text detection and recognition. It leads to the fast oriented text spotting system (FOTS) which can be trained end-to-end. In contrast to previous two-stage text spotting, our method learns more generic features through convolutional neural network, which are shared between text detection and text recognition, and the supervision from the two tasks are complementary. Since feature extraction usually takes most of the time, it shrinks the computation to a single detection network, shown in Fig. \ref{fig:fastdemo}. The key to connect detection and recognition is the \emph{ROIRotate}, which gets proper features from feature maps according to the oriented detection bounding boxes.

The architecture is presented in Fig. \ref{fig:architecture}. Feature maps are firstly extracted with shared convolutions. The fully convolutional network based oriented text detection branch is built on top of the feature map to predict the detection bounding boxes. The RoIRotate operator extracts text proposal features corresponding to the detection results from the feature map. The text proposal features are then fed into Recurrent Neural Network (RNN) encoder and Connectionist Temporal Classification (CTC) decoder \cite{graves2006ctc} for text recognition. Since all the modules in the network are differentiable, the whole system can be trained end-to-end. To the best of our knoweldge, this is the first end-to-end trainable framework for oriented text detection and recognition. We find that the network can be easily trained without complicated post-processing and hyper-parameter tuning.

The contributions are summarized as follows.
\begin{itemize}

\item
We propose an end-to-end trainable framework for fast oriented text spotting. By sharing convolutional features, the network can detect and recognize text simultaneously with little computation overhead, which leads to \emph{real-time} speed.

\item
We introduce the \emph{RoIRotate}, a new differentiable operator to extract the oriented text regions from convolutional feature maps. This operation unifies text detection and recognition into an end-to-end pipeline.

\item
Without bells and whistles, FOTS significantly surpasses state-of-the-art methods on a number of text detection and text spotting benchmarks, including ICDAR 2015 \cite{karatzas2015icdar15}, ICDAR 2017 MLT \cite{icdar17} and ICDAR 2013 \cite{karatzas2013icdar}.

\end{itemize}
\section{Related Work}
Text spotting is an active topic in computer vision and document analysis. In this section, we present a brief introduction to related works including text detection, text recognition and text spotting methods that combine both.

\subsection{Text Detection}
Most conventional methods of text detection consider text as a composition of characters. These character based methods first localize characters in an image and then group them into words or text lines. Sliding-window-based methods \cite{jaderberg2014deep,textbasedapproach,Chen2004Detecting,zhu2016text} and connected-components based methods \cite{huang2013text,neumann2013scene,busta2015fastext} are two representative categories in conventional methods.

Recently, many deep learning based methods are proposed to directly detect words in images. Tian \etal \cite{tian2016ctpn} employ a vertical anchor mechanism to predict the fixed-width sequential proposals and then connect them. Ma \etal \cite{ma2017rcnn} introduce a novel rotation-based framework for arbitrarily oriented text by proposing Rotation RPN and Rotation RoI pooling. Shi \etal \cite{shi2017seglink} first predict text segments and then link them into complete instances using the linkage prediction. With dense predictions and one step post processing, Zhou \etal \cite{zhou2017east} and He \etal \cite{he2017casia} propose deep direct regression methods for multi-oriented scene text detection. 
\subsection{Text Recognition}
Generally, scene text recognition aims to decode a sequence of label from regularly cropped but variable-length text images. Most previous methods \cite{goodfellow2013multi, lee2014region} capture individual characters and refine misclassified characters later. Apart from character level approaches, recent text region recognition approaches can be classified into three categories: word classification based, sequence-to-label decode based and sequence-to-sequence model based methods.

Jaderberg \etal \cite{jaderberg2014synthetic} pose the word recognition problem as a conventional multi-class classification task with a large number of class labels (about 90K words). Su \etal \cite{su2014accurate} frame text recognition as a sequence labelling problem, where RNN is built upon HOG features and adopt CTC as decoder. Shi \etal \cite{shi2016crnn} and He \etal \cite{he2016first} propose deep recurrent models to encode the max-out CNN features and adopt CTC to decode the encoded sequence. Fujii \etal \cite{fujii2017sequence} propose an encoder and summarizer network to produce input sequence for CTC. Lee \etal \cite{lee2016recursive} use an attention-based sequence-to-sequence structure to automatically focus on certain extracted CNN features and implicitly learn a character level language model embodied in RNN. To handle irregular input images, Shi \etal \cite{shi2016robust} and Liu \etal \cite{liu2016star} introduce spatial attention mechanism to transform a distorted text region into a canonical pose suitable for recognition.

\subsection{Text Spotting}
Most previous text spotting methods first generate text proposals using a text detection model and then recognize them with a separate text recognition model. Jaderberg \etal \cite{jaderberg2016ijcv} first generate holistic text proposals with a high recall using an ensemble model, and then use a word classifier for word recognition. Gupta \etal \cite{gupta2016synthetic} train a Fully-Convolutional Regression Network for text detection and adopt the word classifier in \cite{jaderberg2014synthetic} for text recognition. Liao \etal \cite{liao2017textboxes} use an SSD \cite{liu2016ssd} based method for text detection and CRNN \cite{shi2016crnn} for text recognition.

Recently Li \etal \cite{li2017iccv} propose an end-to-end text spotting method, which uses a text proposal network inspired by RPN \cite{ren2015faster} for text detection and LSTM with attention mechanism \cite{luong2015effective,shi2016robust,Chen2004Detecting} for text recognition. Our method has two mainly advantages compared to them: (1) We introduce RoIRotate and use totally different text detection algorithm to solve more complicated and difficult situations, while their method is only suitable for horizontal text. (2) Our method is much better than theirs in terms of speed and performance, and in particular, nearly cost-free text recognition step enables our text spotting system to run at real-time speed, while their method takes approximately 900ms to process an input image of 600$\times$800 pixels.

\section{Methodology}
FOTS is an end-to-end trainable framework that detects and recognizes all words in a natural scene image simultaneously. It consists of four parts: shared convolutions, the text detection branch, RoIRotate operation and the text recognition branch.

\subsection{Overall Architecture}
\label{overall}

\begin{figure}
\includegraphics[width=0.47\textwidth]{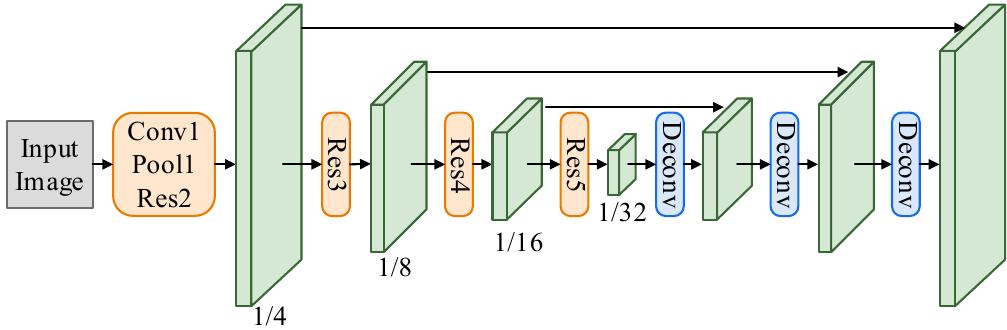}
\vspace{5pt}
\caption{Architecture of shared convolutions. Conv1-Res5 are operations from ResNet-50, and Deconv consists of one convolution to reduce feature channels and one bilinear upsampling operation.}
\label{fig:shared}
\end{figure}

An overview of our framework is illustrated in Fig. \ref{fig:architecture}. The text detection branch and recognition branch share convolutional features, and the architecture of the shared network is shown in Fig. \ref{fig:shared}. The backbone of the shared network is ResNet-50 \cite{he2016resnet}. Inspired by FPN \cite{lin2016fpn}, we concatenate low-level feature maps and high-level semantic feature maps. The resolution of feature maps produced by shared convolutions is 1/4 of the input image. The text detection branch outputs dense per-pixel prediction of text using features produced by shared convolutions. With oriented text region proposals produced by detection branch, the proposed RoIRotate converts corresponding shared features into fixed-height representations while keeping the original region aspect ratio. Finally, the text recognition branch recognizes words in region proposals. CNN and LSTM are adopted to encode text sequence information, followed by a CTC decoder. The structure of our text recognition branch is shown in Tab. \ref{tab:network_detail}.

\begin{table}
\small
\begin{center}
\begin{tabular}{l|l|l}
\hline
\multirow{2}{*}{Type} & Kernel        & \multirow{2}{*}{\begin{tabular}[c]{@{}l@{}}Out\\ Channels\end{tabular}} \\ 
           & {[size, stride]}     &                                   \\ \hline
conv\_bn\_relu    & {[3, 1]}      & 64                                 \\
conv\_bn\_relu    & {[3, 1]}      & 64                                 \\
height-max-pool    & {[(2, 1), (2, 1)]} & 64                                 \\
conv\_bn\_relu    & {[3, 1]}      & 128                                 \\
conv\_bn\_relu    & {[3, 1]}      & 128                                 \\
height-max-pool    & {[(2, 1), (2, 1)]} & 128                                 \\
conv\_bn\_relu    & {[3, 1]}      & 256                                 \\
conv\_bn\_relu    & {[3, 1]}      & 256                                 \\
height-max-pool    & {[(2, 1), (2, 1)]} & 256                                 \\ \hline
bi-directional\_lstm        &           & 256                                 \\
fully-connected    &           & $\vert S\vert$                           \\ \hline
\end{tabular}
\end{center}
\caption{The detailed structure of the text recognition branch. All convolutions are followed by batch normalization and ReLU activation. Note that height-max-pool aims to reduce feature dimension along height axis only.}
\label{tab:network_detail}
\end{table}

\subsection{Text Detection Branch}
\label{detect}
Inspired by \cite{zhou2017east,he2017casia}, we adopt a fully convolutional network as the text detector. As there are a lot of small text boxes in natural scene images, we upscale the feature maps from 1/32 to 1/4 size of the original input image in shared convolutions. After extracting shared features, one convolution is applied to output dense per-pixel predictions of words. The first channel computes the probability of each pixel being a positive sample. Similar to \cite{zhou2017east}, pixels in shrunk version of the original text regions are considered positive. For each positive sample, the following 4 channels predict its distances to top, bottom, left, right sides of the bounding box that contains this pixel, and the last channel predicts the orientation of the related bounding box. Final detection results are produced by applying thresholding and NMS to these positive samples.

In our experiments, we observe that many patterns similar to text strokes are hard to classify, such as fences, lattices, etc. We adopt online hard example mining (OHEM) \cite{shrivastava2016ohem} to better distinguish these patterns, which also solves the class imbalance problem. This provides a F-measure improvement of about 2\% on ICDAR 2015 dataset.

The detection branch loss function is composed of two sterms: text classification term and bounding box regression term. The text classification term can be seen as pixel-wise classification loss for a down-sampled score map. Only shrunk version of the original text region is considered as the positive area, while the area between the bounding box and the shrunk version is considered as ``NOT CARE'', and does not contribute to the loss for the classification. Denote the set of selected positive elements by OHEM in the score map as $\Omega$, the loss function for classification can be formulated as:
\begin{equation}
\begin{split}
L_{\text{cls}} &= \frac{1}{|\Omega|} \sum_{x \in \Omega} \text{H}(p_x, p_x^*) \\
&= \frac{1}{|\Omega|} \sum_{x \in \Omega} (-p_x^*\log p_x - (1 - p_x^*)\log(1 - p_x))
\end{split}
\end{equation}
where $|\cdot|$ is the number of elements in a set, and H$(p_x, p_x^*)$ represents the cross entropy loss between $p_x$, the prediction of the score map, and $p_x^*$, the binary label that indicates text or non-text.

As for the regression loss, we adopt the IoU loss in \cite{yu2016unitbox} and the rotation angle loss in \cite{zhou2017east}, since they are robust to variation in object shape, scale and orientation:
\begin{equation}
L_{\text{reg}} = \frac{1}{|\Omega|} \sum_{x \in \Omega} \text{IoU}(\mathbf{R}_x, \mathbf{R}_x^*) + \lambda_{\theta}(1-\cos(\theta_x, \theta_x^*)) \label{reg}
\end{equation}
Here, $\text{IoU}(\mathbf{R}_x, \mathbf{R}_x^*)$ is the IoU loss between the predicted bounding box $\mathbf{R}_x$, and the ground truth $\mathbf{R}_x^*$. The second term is rotation angle loss, where $\theta_x$ and $\theta_x^*$ represent predicted orientation and the ground truth orientation respectively. We set the hyper-parameter $\lambda_{\theta}$ to 10 in experiments.

Therefore the full detection loss can be written as:
\begin{equation}
L_{\text{detect}} = L_{\text{cls}} + \lambda_{\text{reg}}L_{\text{reg}} \label{L_detect}
\end{equation}
where a hyper-parameter $\lambda_{\text{reg}}$ balances two losses, which is set to 1 in our experiments.

\subsection{RoIRotate}
\label{tpl}

\begin{figure}
\centering
\includegraphics[width=0.45\textwidth]{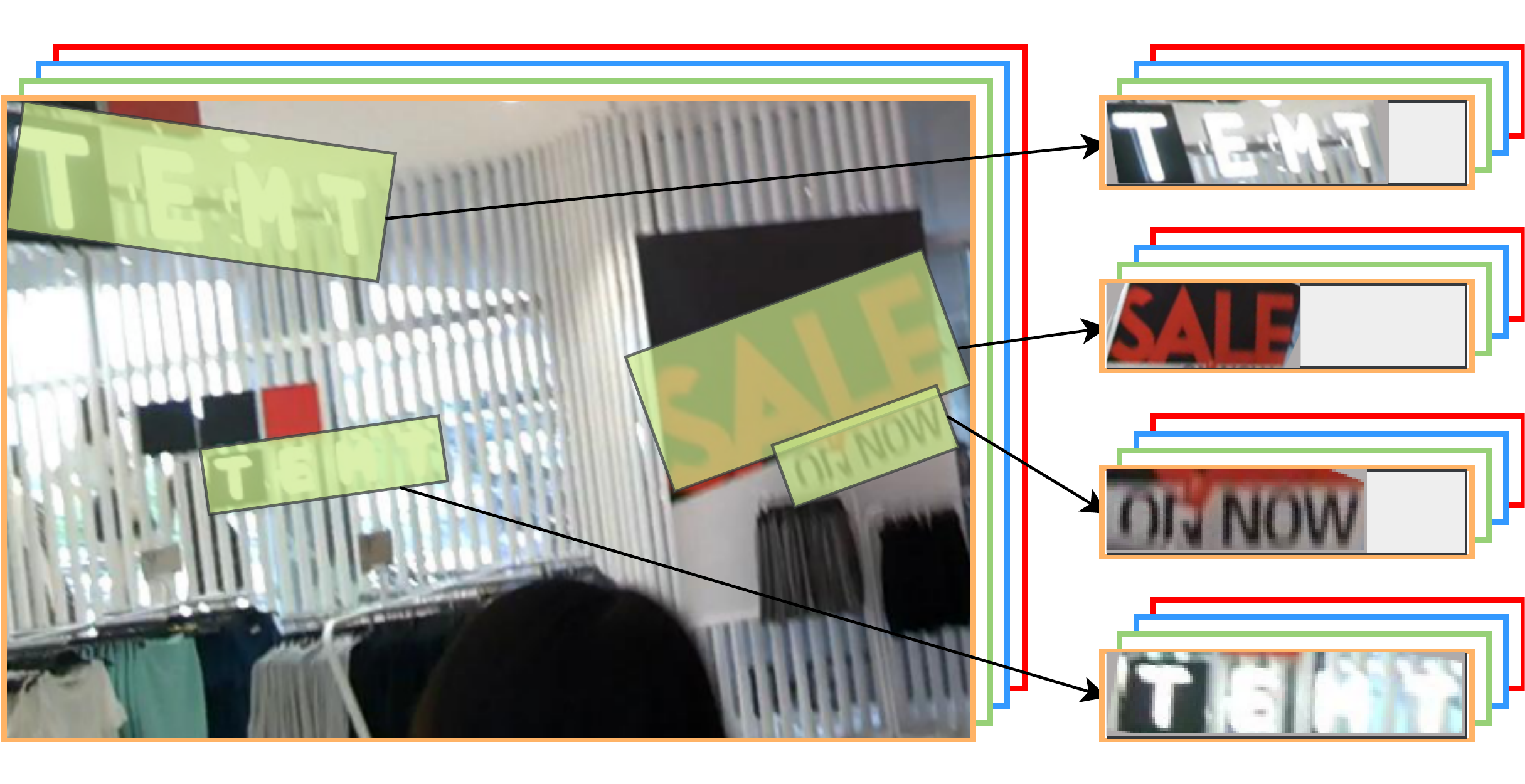}
\vspace{5pt}
\caption{Illustration of RoIRotate. Here we use the input image to illustrate text locations, but it is actually operated on feature maps in the network. Best view in color.}
\label{fig:tpl}
\end{figure}

RoIRotate applies transformation on oriented feature regions to obtain axis-aligned feature maps, as shown in Fig. \ref{fig:tpl}. In this work, we fix the output height and keep the aspect ratio unchanged to deal with the variation in text length. Compared to RoI pooling \cite{girshick2015fast} and RoIAlign \cite{he2017mask}, RoIRotate provides a more general operation for extracting features for regions of interest. We also compare to RRoI pooling proposed in RRPN \cite{ma2017rcnn}. RRoI pooling transforms the rotated region to a fixed size region through max-pooling, while we use bilinear interpolation to compute the values of the output. This operation avoids misalignments between the RoI and the extracted features, and additionally it makes the lengths of the output features variable, which is more suitable for text recognition.

This process can be divided into two steps. First, affine transformation parameters are computed via predicted or ground truth coordinates of text proposals. Then, affine transformations are applied to shared feature maps for each region respectively, and canonical horizontal feature maps of text regions are obtained. The first step can be formulated as:
{
\setlength\abovedisplayskip{0pt}
\setlength\belowdisplayskip{0pt}
\begin{align}
t_x &= l * \cos \theta - t *  \sin \theta - x \\
t_y &= t * \cos \theta + l *  \sin \theta - y \\
s &= \frac{h_t}{t+b} \\
w_t &= s * (l+r)
\end{align}
}
{
\setlength\abovedisplayskip{0pt}
\setlength\belowdisplayskip{1em}
\begin{align}
\mathbf{M} &= \begin{bmatrix}
\cos\theta & -\sin \theta & 0 \\ \sin \theta & \cos \theta &0 \\ 0 & 0 & 1
\end{bmatrix}\begin{bmatrix}
s & 0 & 0 \\ 0 & s & 0 \\ 0 & 0 & 1
\end{bmatrix}
\begin{bmatrix}
1 & 0 & t_x \\ 0 & 1 & t_y \\ 0 & 0 & 1
\end{bmatrix} \notag \\
&= s \begin{bmatrix}
\cos \theta & -\sin \theta & t_x \cos \theta - t_y \sin \theta \\
\sin \theta & \cos \theta & t_x \sin \theta + t_y \cos \theta \\
0 & 0 & \frac{1}{s}
\end{bmatrix}
\end{align}
}
\newline
where $\mathbf{M}$ is the affine transformation matrix. $h_t, w_t$ represent height (equals 8 in our setting) and width of feature maps after affine transformation. $(x, y)$ represents the coordinates of a point in shared feature maps and $(t, b, l, r)$ stands for distance to top, bottom, left, right sides of the text proposal respectively, and $\theta$ for the orientation. $(t, b, l, r)$ and $\theta$ can be given by ground truth or the detection branch.

With the transformation parameters, it is easy to produce the final RoI feature using the affine transformation:
\begin{equation}
\begin{pmatrix}
x_i^s \\ y_i^s \\1
\end{pmatrix} = \mathbf{M}^{-1} \begin{pmatrix}
x_i^t \\ y_i^t \\1
\end{pmatrix}
\end{equation}
and for $\forall i \in [1\ldots h_t]$, $\forall j \in [1...w_t]$, $\forall c \in [1\ldots C]$,
\begin{equation}
V_{ij}^c = \sum_{n}^{h_s} \sum_{m}^{w_s} U_{nm}^{c} k(x_{ij}^s -m;\Phi_x) k(y_{ij}^s-n;\Phi_y)
\end{equation}
where $V_{ij}^c$ is the output value at location $(i, j)$ in channel $c$ and $U_{nm}^{c}$ is the input value at location $(n, m)$ in channel $c$. $h_s$, $w_s$ represent the height and width of the input, and $\Phi_x$, $\Phi_y$ are the parameters of a generic sampling kernel $k()$, which defines the interpolation method, specifically bilinear interpolation in this work. As the width of text proposals may vary, in practice, we pad the feature maps to the longest width and ignore the padding parts in recognition loss function.

Spatial transformer network \cite{jaderberg2015stn} uses affine transformation in a similar way, but gets transformation parameters via a different method and is mainly used in the image domain, i.e. transforming images themselves. RoIRotate takes feature maps produced by shared convolutions as input, and generates the feature maps of all text proposals, with fixed height and unchanged aspect ratio.

Different from object classification, text recognition is very sensitive to detection noise. A small error in predicted text region could cut off several characters, which is harmful to network training, so we use ground truth text regions instead of predicted text regions during training. When testing, thresholding and NMS are applied to filter predicted text regions. After RoIRotate, transformed feature maps are fed to the text recognition branch.

\subsection{Text Recognition Branch}
\label{recog}
The text recognition branch aims to predict text labels using the region features extracted by shared convolutions and transformed by RoIRotate. Considering the length of the label sequence in text regions, input features to LSTM are reduced only twice (to 1/4 as described in Sec. \ref{detect}) along width axis through shared convolutions from the original image. Otherwise discriminable features in compact text regions, especially those of narrow shaped characters, will be eliminated. Our text recognition branch consists of VGG-like \cite{simonyan2014vgg} sequential convolutions, poolings with reduction along height axis only, one bi-directional LSTM \cite{schuster1997bidirectional,hochreiter1997lstm}, one fully-connection and the final CTC decoder \cite{graves2006ctc}.

First, spatial features are fed into several sequential convolutions and poolings along height axis with dimension reduction to extract higher-level features. For simplicity, all reported results here are based on VGG-like sequential layers as shown in Tab. \ref{tab:network_detail}.

Next, the extracted higher-level feature maps $\mathbf{L} \in \mathbb{R}^{C \times H \times W}$ are permuted to time major form as a sequence $ \bm{l}_1, ..., \bm{l}_W \in \mathbb{R}^{C \times H}$ and fed into RNN for encoding. Here we use a bi-directional LSTM, with $D=256$ output channels per direction, to capture range dependencies of the input sequential features. Then, hidden states $\bm{h}_1, ..., \bm{h}_W \in \mathbb{R}^{D}$ calculated at each time step in both directions are summed up and fed into a fully-connection, which gives each state its distribution $\bm{x}_t \in \mathbb{R}^{\vert S \vert}$ over the character classes $S$. To avoid overfitting on small training datasets like ICDAR 2015, we add dropout before fully-connection. Finally, CTC is used to transform frame-wise classification scores to label sequence. Given probability distribution $\bm{x}_t$ over $S$ of each $\bm{h}_t $, and ground truth label sequence $\bm{y}^* = \{y_1, ..., y_T \}, T \leqslant W$, the conditional probability of the label $\bm{y}^*$ is the sum of probabilities of all paths $\pi$ agreeing with \cite{graves2006ctc}:
\begin{equation}
p(\bm{y}^* \vert \bm{x}) = \sum_{\pi \in {\cal B}^{-1}(\bm{y}^*)} p(\pi \vert \bm{x}) \label{cond_prob}
\end{equation}
where ${\cal B}$ defines a many-to-one map from the set of possible labellings with blanks and repeated labels to $\bm{y}^*$. The training process attempts to maximize the log likelihood of summation of Eq. \eqref{cond_prob} over the whole training set. Following \cite{graves2006ctc}, the recognition loss can be formulated as:
\begin{equation}
L_{\text{recog}} = -\frac{1}{N} \sum_{n=1}^{N} \log p(\bm{y}_n^* \vert \bm{x})
\end{equation}
where $N$ is the number of text regions in an input image, and $\bm{y}_n^*$ is the recognition label.

Combined with detection loss $L_{\text{detect}}$ in Eq. \eqref{L_detect}, the full multi-task loss function is:
\begin{equation}
L = L_{\text{detect}} + \lambda_{\text{recog}} L_{\text{recog}}
\end{equation}
where a hyper-parameter $\lambda_{\text{recog}}$ controls the trade-off between two losses. $\lambda_{\text{recog}}$ is set to 1 in our experiments.
\begin{table*}
\small
\begin{center}
\setlength{\tabcolsep}{5pt}
\begin{tabular}{l|ccc|l|ccc|ccc}
\hline
\multirow{2}{*}{Method} & \multicolumn{3}{c|}{Detection} & \multirow{2}{*}{Method} & \multicolumn{3}{c|}{End-to-End} & \multicolumn{3}{c}{Word Spotting}\\ \cline{2-4} \cline{6-11}
 & P & R & F & & S & W & G & S & W & G \\ \hline
SegLink \cite{shi2017seglink} & 74.74 & 76.50 & 75.61 & Baseline OpenCV3.0+Tesseract \cite{karatzas2015icdar15} & 13.84 & 12.01 & 8.01 & 14.65 & 12.63 & 8.43 \\
SSTD \cite{he2017sstd} & 80.23 & 73.86 & 76.91 & Deep2Text-MO \cite{yin2014robust,yin2015multi,jaderberg2016ijcv} & 16.77 & 16.77 & 16.77 & 17.58 & 17.58 & 17.58\\
WordSup \cite{hu2017wordsup} & 79.33 & 77.03& 78.16 & Beam search CUNI+S \cite{karatzas2015icdar15} & 22.14 & 19.80 & 17.46 & 23.37 & 21.07 & 18.38\\
RRPN \cite{ma2017rcnn} & 83.52 & 77.13 & 80.20 & NJU Text (Version3) \cite{karatzas2015icdar15} & 32.63 & - & - & 34.10 & - & - \\
EAST \cite{zhou2017east} & 83.27 & 78.33 & 80.72 & StradVision\_v1 \cite{karatzas2015icdar15} & 33.21 & - & - & 34.65 & - & - \\
NLPR-CASIA \cite{he2017casia} & 82 & 80 & 81 & Stradvision-2 \cite{karatzas2015icdar15} & 43.70 & - & - & 45.87 & - & - \\
R$^2$CNN \cite{jiang2017r2cnn} & 85.62 &79.68 & 82.54 & TextProposals+DictNet \cite{gomez2017textproposals,jaderberg2014synthetic} & 53.30 & 49.61 & 47.18 & 56.00 & 52.26 & 49.73 \\
CCFLAB\_FTSN \cite{dai2017fused} & 88.65 &80.07 &84.14 & HUST\_MCLAB \cite{shi2017seglink,shi2016crnn} & 67.86 & - & - & 70.57 & - & - \\ \hline
Our Detection & 88.84 & 82.04& 85.31 & Our Two-Stage & 77.11 & 74.54 &58.36 &80.38 &77.66 & 58.19\\
FOTS & 91.0 &	85.17 &	87.99 & FOTS & 81.09 & 75.90 & 60.80 & 84.68 & 79.32 & 63.29 \\
FOTS RT & 85.95 & 79.83&	82.78 & FOTS RT & 73.45 & 66.31 & 51.40 &76.74 &69.23 &53.50 \\
FOTS MS &\textbf{91.85}&\textbf{87.92}&\textbf{89.84}& FOTS MS &\textbf{83.55}&\textbf{79.11}&\textbf{65.33}&\textbf{87.01}&\textbf{82.39}&\textbf{67.97} \\ \hline
\end{tabular}
\end{center}
\caption{Comparison with other results on ICDAR 2015 with percentage scores. ``FOTS MS'' represents multi-scale testing and ``FOTS RT'' represents our real-time version, which will be discussed in Sec. \ref{speed}. ``End-to-End'' and ``Word Spotting'' are two types of evaluation protocols for text spotting. ``P'', ``R'', ``F'' represent ``Precision'', ``Recall'', ``F-measure'' respectively and ``S'', ``W'', ``G'' represent F-measure using ``Strong'', ``Weak'', ``Generic'' lexicon respectively.}
\label{tab:icdar15_detect_compare}
\end{table*}

\begin{table}
\small
\begin{center}
\begin{tabular}{l|ccc}
\hline
Method    & Precision   & Recall   & F-measure \\ \hline
linkage-ER-Flow \cite{icdar17} & 44.48 & 25.59 & 32.49   \\
TH-DL \cite{icdar17} & 67.75 &	34.78 &	45.97    \\
TDN\_SJTU2017 \cite{icdar17} & 64.27 &	47.13 &	54.38  \\
SARI\_FDU\_RRPN\_v1 \cite{ma2017rcnn} & 71.17 & 55.50 &	62.37   \\
SCUT\_DLVClab1 \cite{icdar17} & 80.28 &	54.54 &	64.96   \\ \hline
Our Detection & 79.48&	57.45&	66.69\\
FOTS & 80.95&	57.51&	67.25    \\
FOTS MS &\textbf{81.86}&\textbf{62.30}&	\textbf{70.75} \\ \hline
\end{tabular}
\end{center}
\caption{Comparison with other results on ICDAR 2017 MLT scene text detection task.}
\label{tab:MLT_detect_compare}
\end{table}

\begin{table*}
\small
\begin{center}
\begin{tabular}{l|cc|l|ccc|ccc}
\hline
\multirow{2}{*}{Method} & \multicolumn{2}{c|}{Detection} & \multirow{2}{*}{Method} & \multicolumn{3}{c|}{End-to-End} & \multicolumn{3}{c}{Word Spotting}\\ \cline{2-3} \cline{5-10}
 & IC13 & DetEval & & S & W & G & S & W & G \\ \hline
TextBoxes \cite{liao2017textboxes} & 85& 86 &NJU Text (Version3) \cite{karatzas2013icdar} & 74.42 &-&-&77.89&-&-\\
CTPN \cite{tian2016ctpn} & 82.15 & 87.69&StradVision-1 \cite{karatzas2013icdar} &81.28&78.51&67.15&85.82&82.84&70.19\\
R$^2$CNN \cite{jiang2017r2cnn} &79.68 & 87.73 & Deep2Text II+ \cite{yin2014robust,jaderberg2016ijcv} & 81.81&79.47&76.99 & 84.84&83.43&78.90\\
NLPR-CASIA \cite{he2017casia} & 86 & - & VGGMaxBBNet(055) \cite{jaderberg2016ijcv,jaderberg2014synthetic} & 86.35 & - & - & 90.49 & - & 76 \\
SSTD \cite{he2017sstd} & 87 & 88 & FCRNall+multi-filt \cite{gupta2016synthetic} & - & - &- &- & - & 84.7\\
WordSup \cite{hu2017wordsup} & - & 90.34 & Adelaide\_ConvLSTMs \cite{li2016reading} & 87.19 & 86.39 & 80.12 & 91.39 & 90.16 & 82.91 \\
RRPN \cite{ma2017rcnn} & - & 91 & TextBoxes \cite{liao2017textboxes} & 91.57&89.65&83.89& 93.90&91.95&85.92\\
Jiang \etal \cite{jiang2017deep} & 89.54 & 91.85 & Li \etal \cite{li2017iccv} & 91.08&89.81&84.59& 94.16 &92.42&\textbf{88.20} \\ \hline
Our Detection & 86.96 & 87.32 & Our Two-Stage& 87.84 & 86.96& 80.79&91.70 &90.68 & 82.97 \\
FOTS & 88.23 & 88.30 & FOTS & 88.81& 87.11 & 80.81 & 92.73 & 90.72 & 83.51 \\
FOTS MS &\textbf{92.50}&\textbf{92.82}& FOTS MS &\textbf{91.99}&\textbf{90.11}&\textbf{84.77}&\textbf{95.94}&\textbf{93.90}& 87.76 \\ \hline
\end{tabular}
\end{center}
\caption{Comparison with other results on ICDAR 2013. ``IC03'' and ``DetEval'' represent F-measure under ICDAR 2013 evaluation and DetEval evaluation respectively.}
\label{tab:icdar13_detect_compare}
\end{table*}

\subsection{Implementation Details}
\label{train}
We use model trained on ImageNet dataset \cite{krizhevsky2012imagenet} as our pre-trained model. The training process includes two steps: first we use Synth800k dataset \cite{gupta2016synthetic} to train the network for 10 epochs, and then real data is adopted to fine-tune the model until convergence. Different training datasets are adopted for different tasks, which will be discussed in Sec. \ref{exper}. Some blurred text regions in ICDAR 2015 and ICDAR 2017 MLT datasets are labeled as ``DO NOT CARE'', and we ignore them in training.

Data augmentation is important for robustness of deep neural networks, especially when the number of real data is limited, as in our case. First, longer sides of images are resized from 640 pixels to 2560 pixels. Next, images are rotated in range $ [-10^\circ, 10^\circ] $ randomly. Then, the heights of images are rescaled with ratio from 0.8 to 1.2 while their widths keep unchanged. Finally, 640$\times$640 random samples are cropped from the transformed images.

As described in Sec. \ref{detect}, we adopt OHEM for better performance. For each image, 512 hard negative samples, 512 random negative samples and all positive samples are selected for classification. As a result, positive-to-negative ratio is increased from 1:60 to 1:3. And for bounding box regression, we select 128 hard positive samples and 128 random positive samples from each image for training.

At test time, after getting predicted text regions from the text detection branch, the proposed RoIRotate applys thresholding and NMS to these text regions and feeds selected text features to the text recognition branch to get final recognition result. For multi-scale testing, results from all scales are combined and fed to NMS again to get the final results.
\section{Experiments}
\label{exper}

We evaluate the proposed method on three recent challenging public benchmarks: ICDAR 2015 \cite{karatzas2015icdar15}, ICDAR 2017 MLT \cite{icdar17} and ICDAR 2013 \cite{karatzas2013icdar}, and surpasses state-of-the-art methods in both text localization and text spotting tasks. All the training data we use is publicly available.

\subsection{Benchmark Datasets}

\textbf{ICDAR 2015} is the Challenge 4 of ICDAR 2015 Robust Reading Competition, which is commonly used for oriented scene text detection and spotting. This dataset includes 1000 training images and 500 testing images. These images are captured by Google glasses without taking care of position, so text in the scene can be in arbitrary orientations. For text spotting task, it provides 3 specific lists of words as lexicons for reference in the test phase, named as ``Strong'', ``Weak'' and ``Generic''. ``Strong'' lexicon provides 100 words per-image including all words that appear in the image. ``Weak'' lexicon includes all words that appear in the entire test set. And ``Generic'' lexicon is a 90k word vocabulary. In training, we first train our model using 9000 images from ICDAR 2017 MLT training and validation datasets, then we use 1000 ICDAR 2015 training images and 229 ICDAR 2013 training images to fine-tune our model.

\textbf{ICDAR 2017 MLT} is a large scale multi-lingual text dataset, which includes 7200 training images, 1800 validation images and 9000 testing images. The dataset is composed of complete scene images which come from 9 languages, and text regions in this dataset can be in arbitrary orientations, so it is more diverse and challenging. This dataset does not have text spotting task so we only report our text detection result. We use both training set and validation set to train our model.

\textbf{ICDAR 2013} consists of 229 training images and 233 testing images, and similar to ICDAR 2015, it also provides ``Strong'', ``Weak'' and ``Generic'' lexicons for text spotting task. Different to above datasets, it contains only horizontal text. Though our method is designed for oriented text, results in this dataset indicate the proposed method is also suitable for horizontal text. Due to there are too few training images, we first use 9000 images from ICDAR 2017 MLT training and validation datasets to train a pre-trained model and then use 229 ICDAR 2013 training images to fine-tune.

\begin{figure*}[t]
\begin{center}
\hspace{-5pt}
\begin{minipage}[b]{9pt}
 \small
 \vspace{10pt}
 \rotatebox{90}{Our Detection\hspace{30pt}FOTS}
 \vspace{6pt}
\end{minipage}
\hspace{-5pt}
\subfigure[Miss]{
\begin{minipage}[b]{0.235\textwidth}
\includegraphics[width=\textwidth]{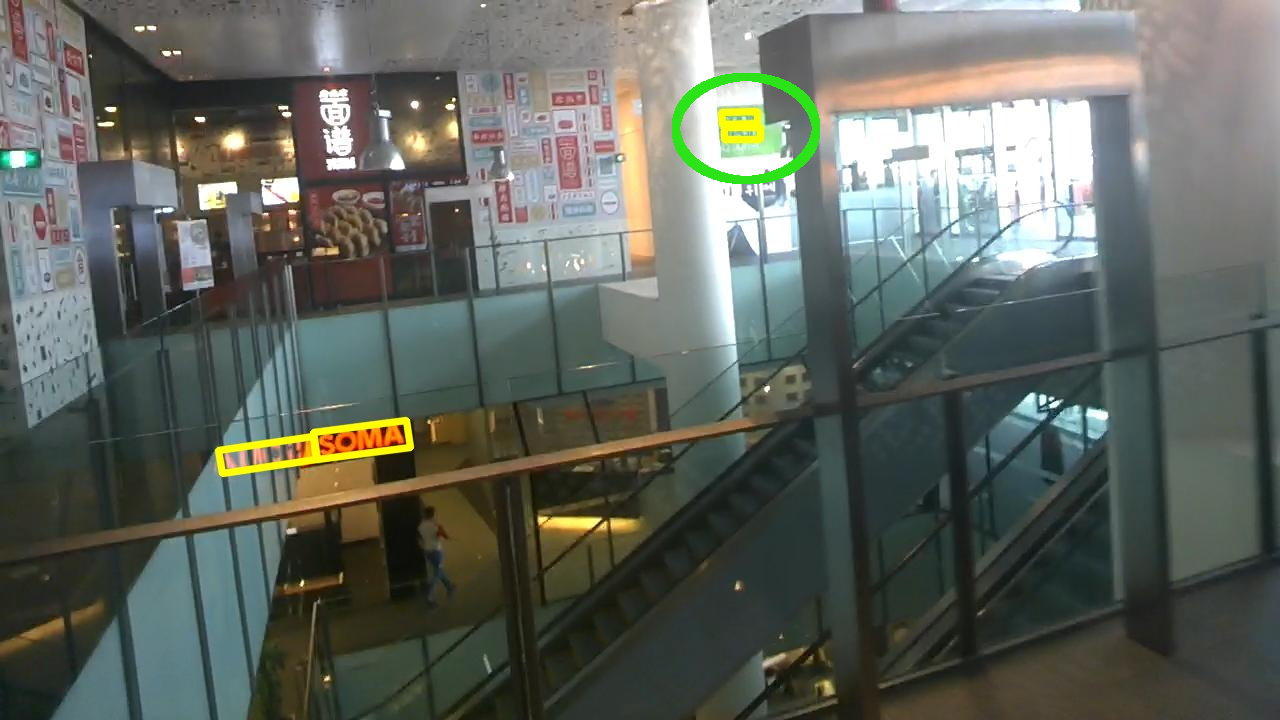} \\
\vspace{-10pt}
\includegraphics[width=\textwidth]{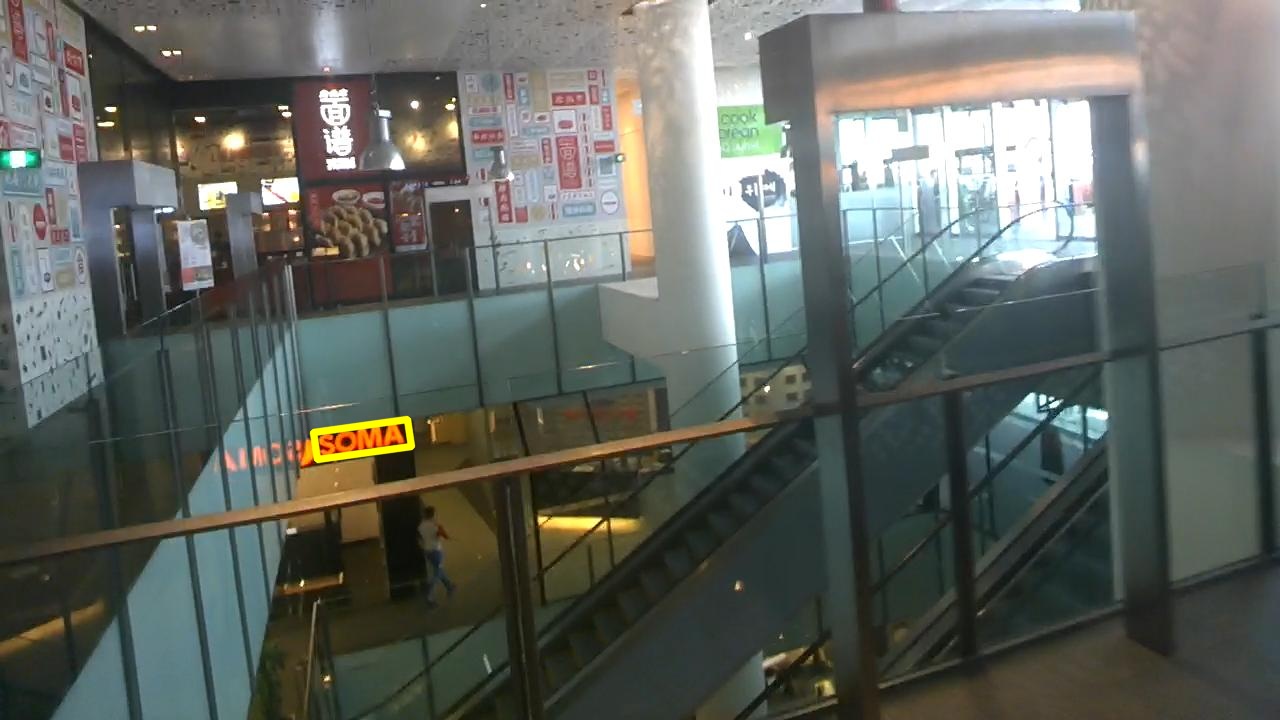}
\end{minipage}
}\hspace{-0.07in}
\subfigure[False]{
\begin{minipage}[b]{0.235\textwidth}
\includegraphics[width=\textwidth]{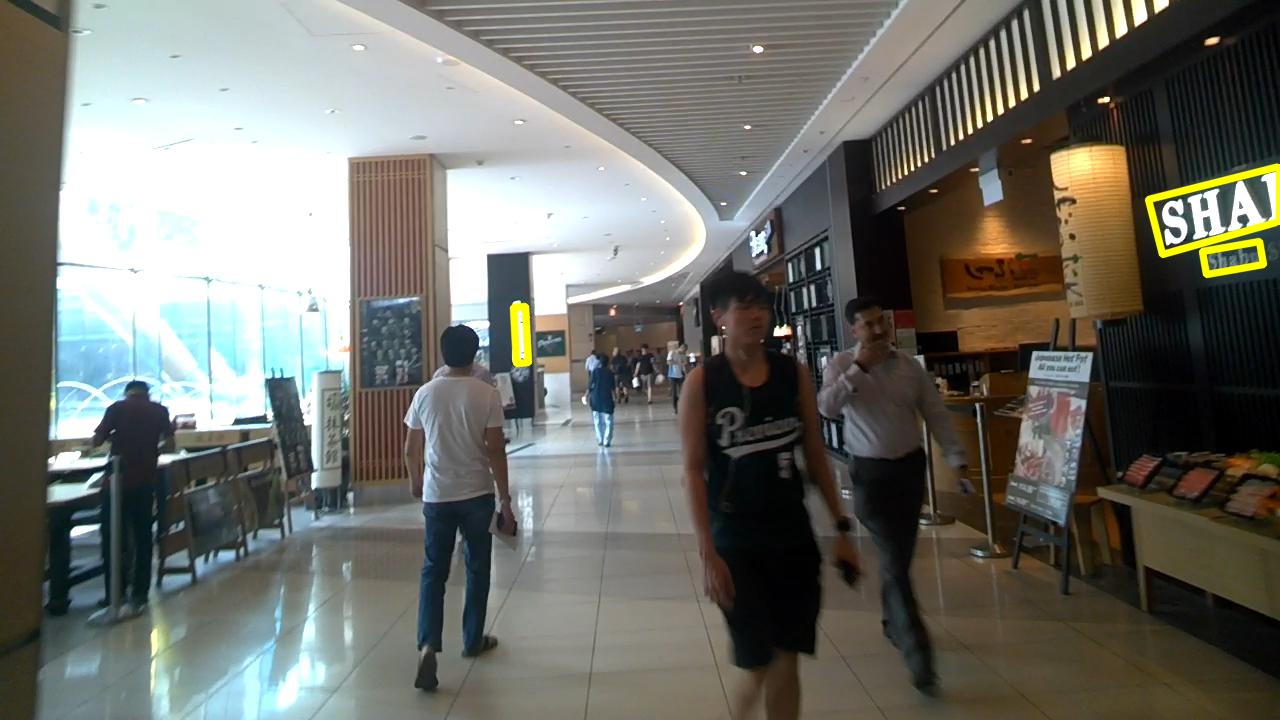} \\
\vspace{-10pt}
\includegraphics[width=\textwidth]{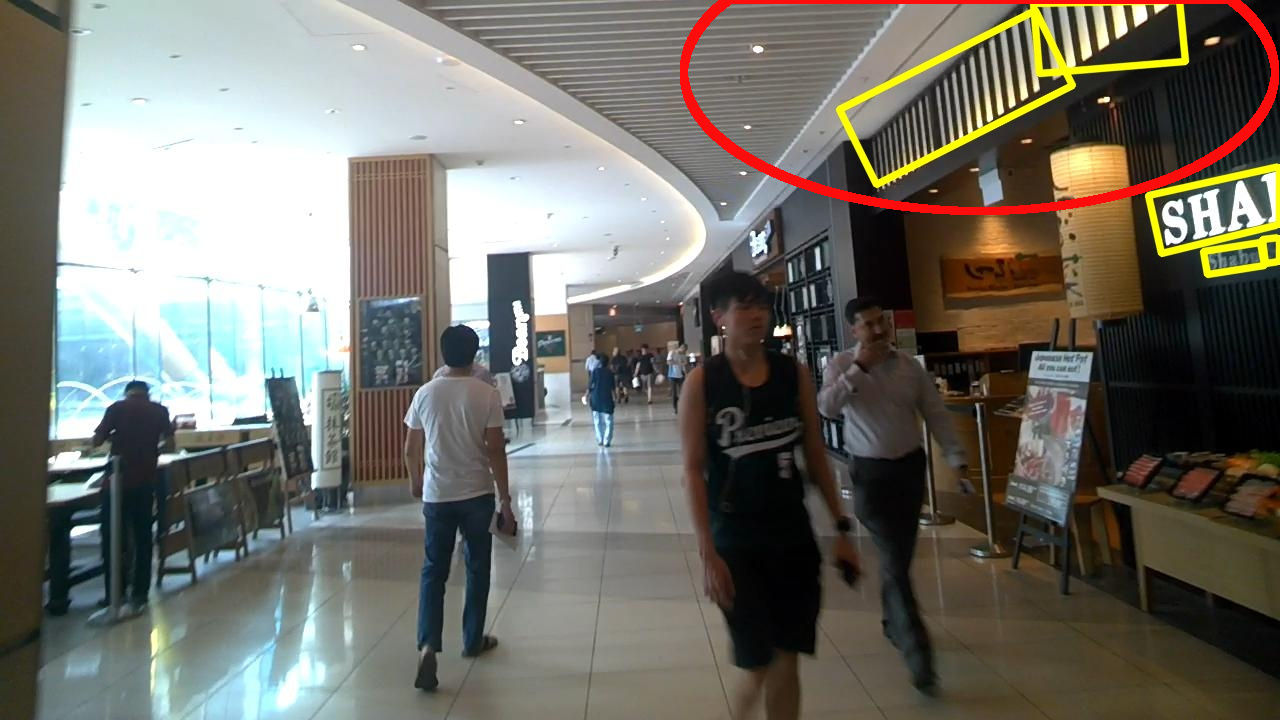}
\end{minipage}
}\hspace{-0.07in}
\subfigure[Split]{
\begin{minipage}[b]{0.235\textwidth}
\includegraphics[width=\textwidth]{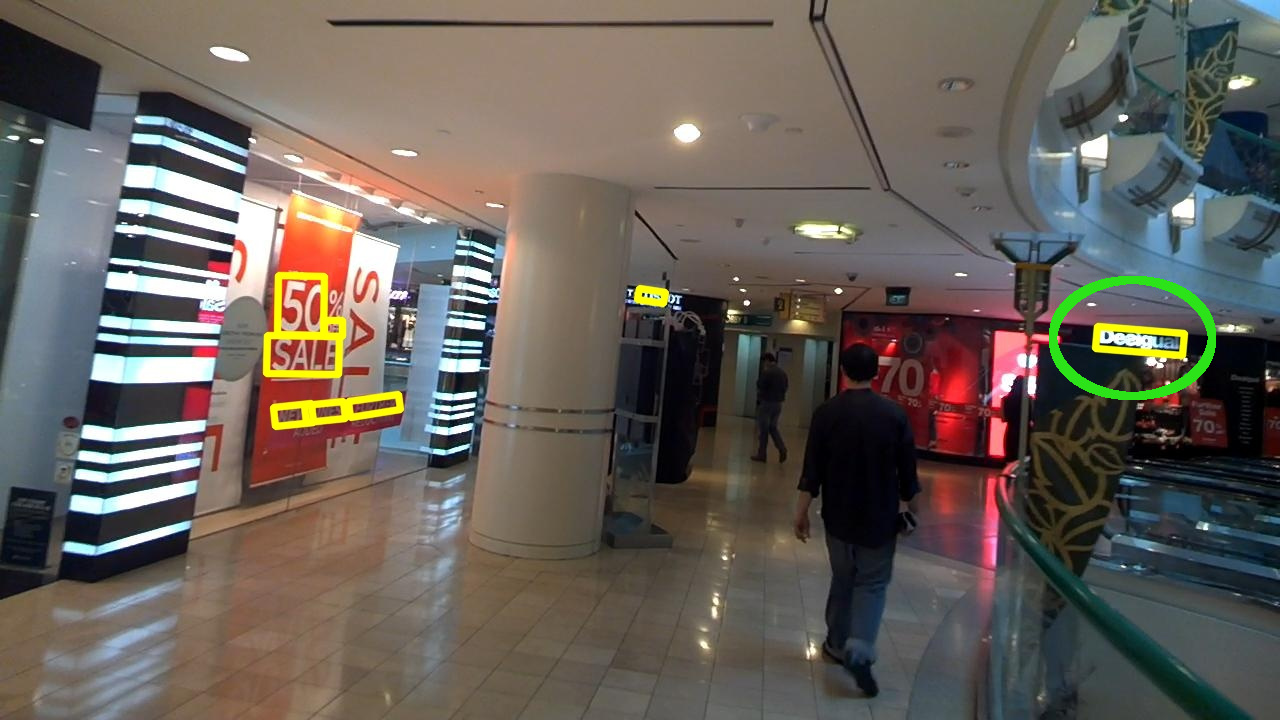} \\
\vspace{-10pt}
\includegraphics[width=\textwidth]{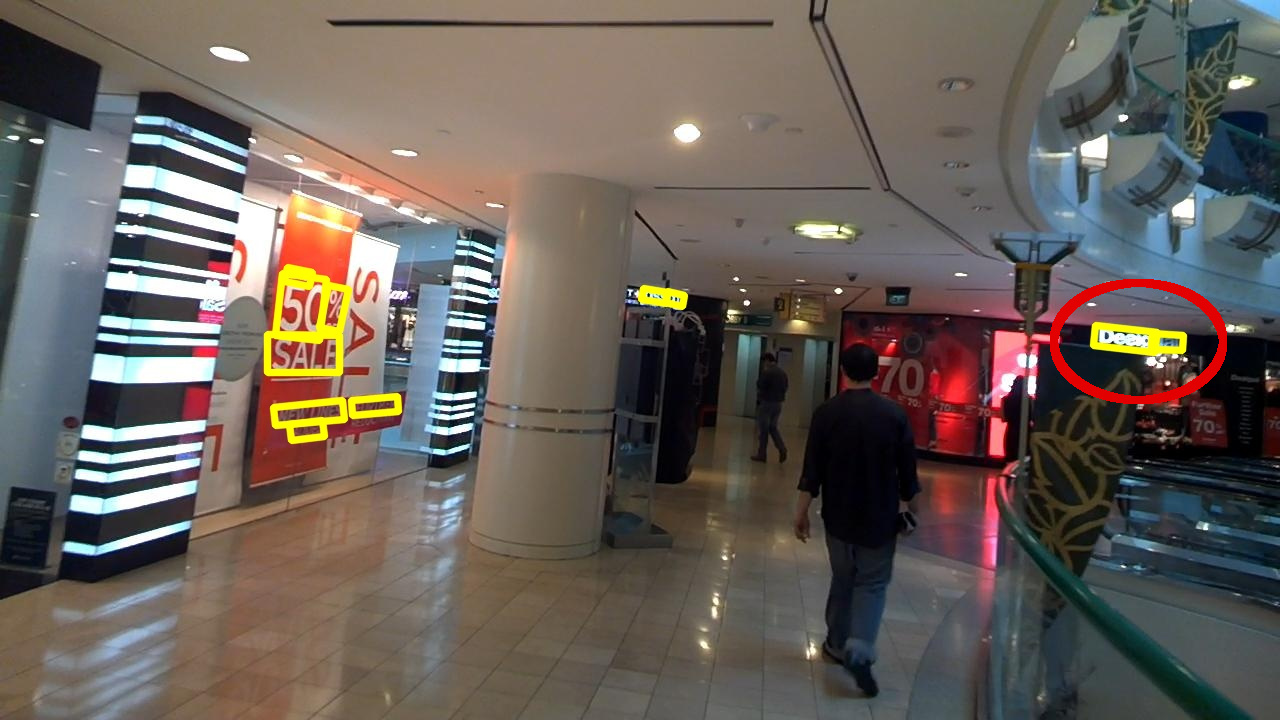}
\end{minipage}
}\hspace{-0.07in}
\subfigure[Merge]{
\begin{minipage}[b]{0.235\textwidth}
\includegraphics[width=\textwidth]{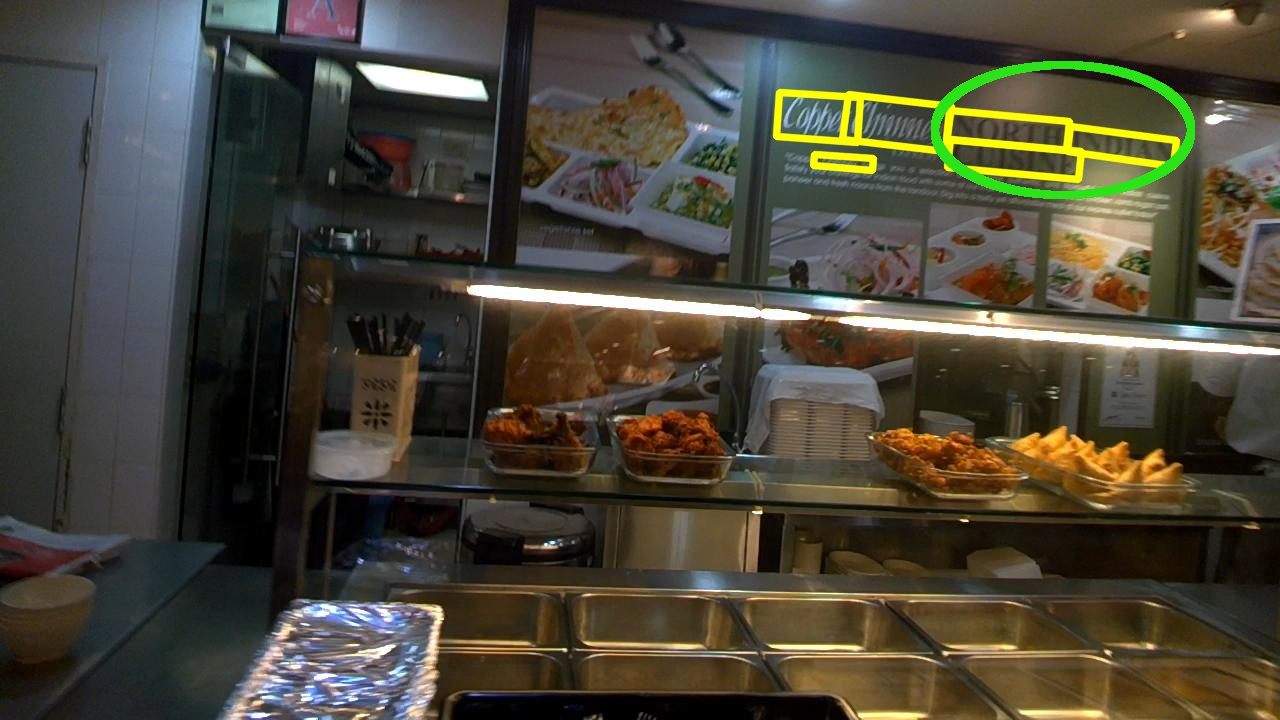} \\
\vspace{-10pt}
\includegraphics[width=\textwidth]{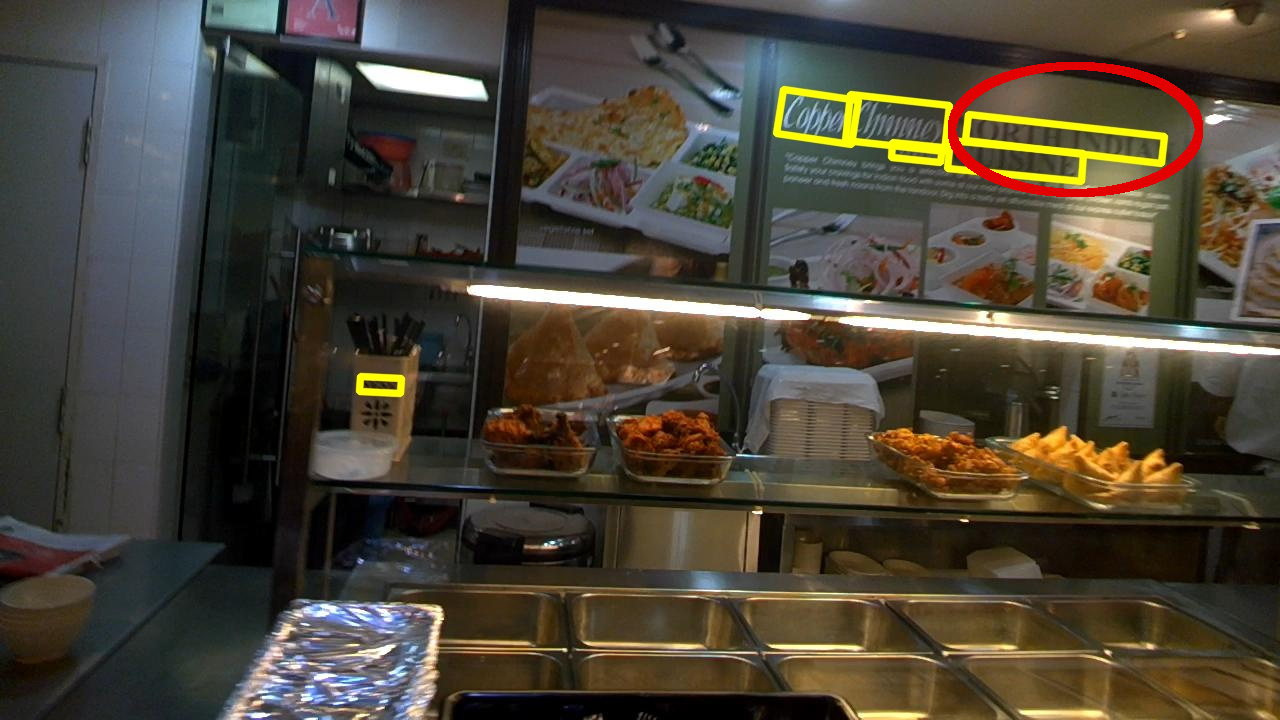}
\end{minipage}
}
\end{center}
\vspace{-10pt}
\caption{FOTS reduces Miss, False, Split and Merge errors in detection. Bounding boxes in green ellipses represent correct text regions detected by FOTS, and those in red ellipses represent wrong text regions detected by ``Our Detection" method. Best view in color.}
\label{tab:example}
\end{figure*}

\subsection{Comparison with Two-Stage Method}
\label{e2e_vs_2stage}
Different from previous works which divide text detection and recognition into two unrelated tasks, our method train these two tasks jointly, and both text detection and recognition can benefit from each other. To verify this, we build a two-stage system, in which text detection and recognition models are trained separately. The detection network is built by removing recognition branch in our proposed network, and similarly, detection branch is removed from origin network to get the recognition network. For recognition network, text line regions cropped from source images are used as training data, similar to previous text recognition methods \cite{shi2016crnn, he2016first, liu2016star}.

As shown in Tab. \ref{tab:icdar15_detect_compare},\ref{tab:MLT_detect_compare},\ref{tab:icdar13_detect_compare}, our proposed FOTS significantly outperforms the two-stage method ``Our Detection" in text localization task and ``Our Two-Stage" in text spotting task. Results show that our joint training strategy pushes model parameters to a better converged state.

\begin{figure*}[!ht]
\begin{center}
\subfigure[ICDAR 2015]{
\begin{minipage}[b]{0.3\textwidth}
\includegraphics[width=\textwidth]{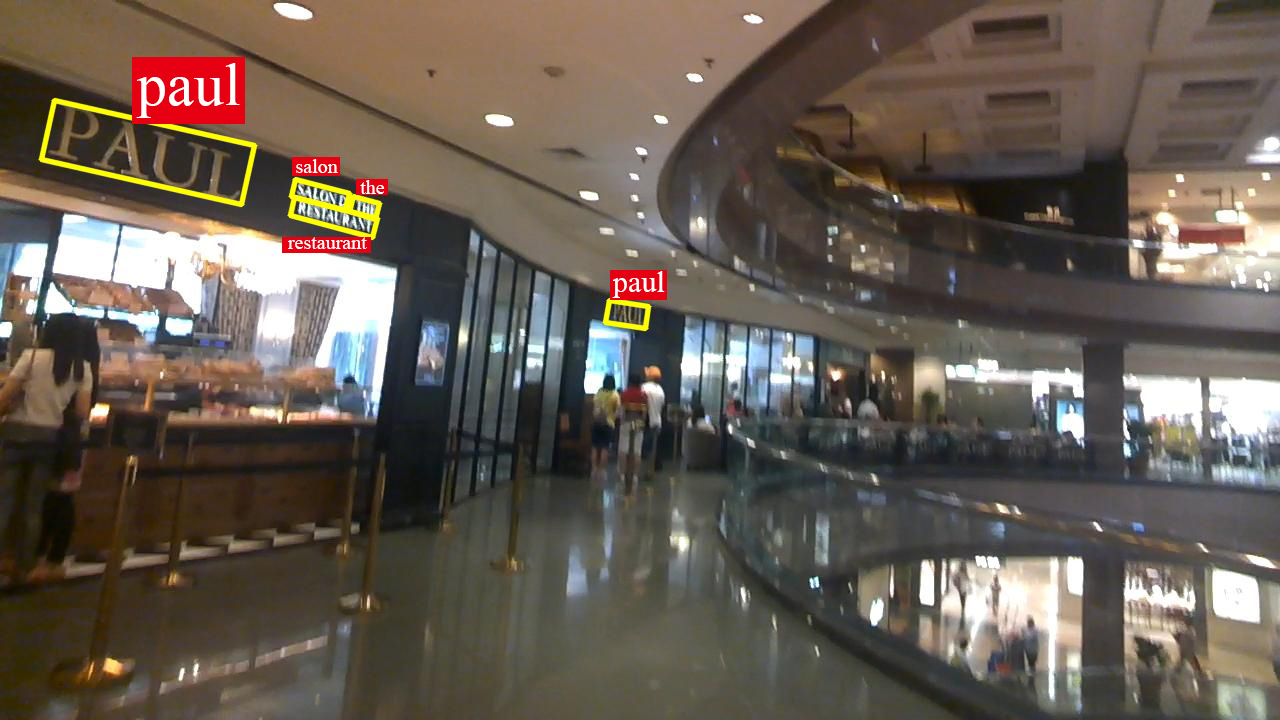} \vspace{-10pt}\\
\includegraphics[width=\textwidth]{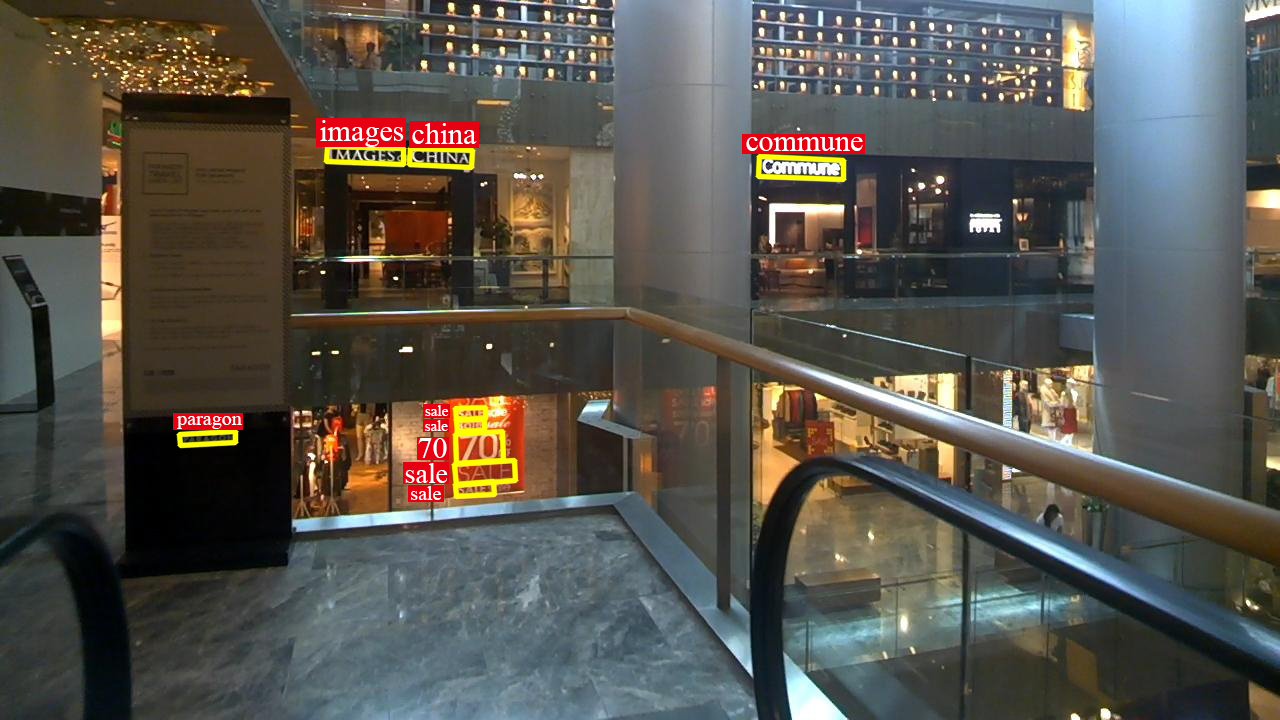}
\end{minipage}
}
\subfigure[ICDAR 2017 MLT]{
\begin{minipage}[b]{0.222\textwidth}
\includegraphics[width=\textwidth]{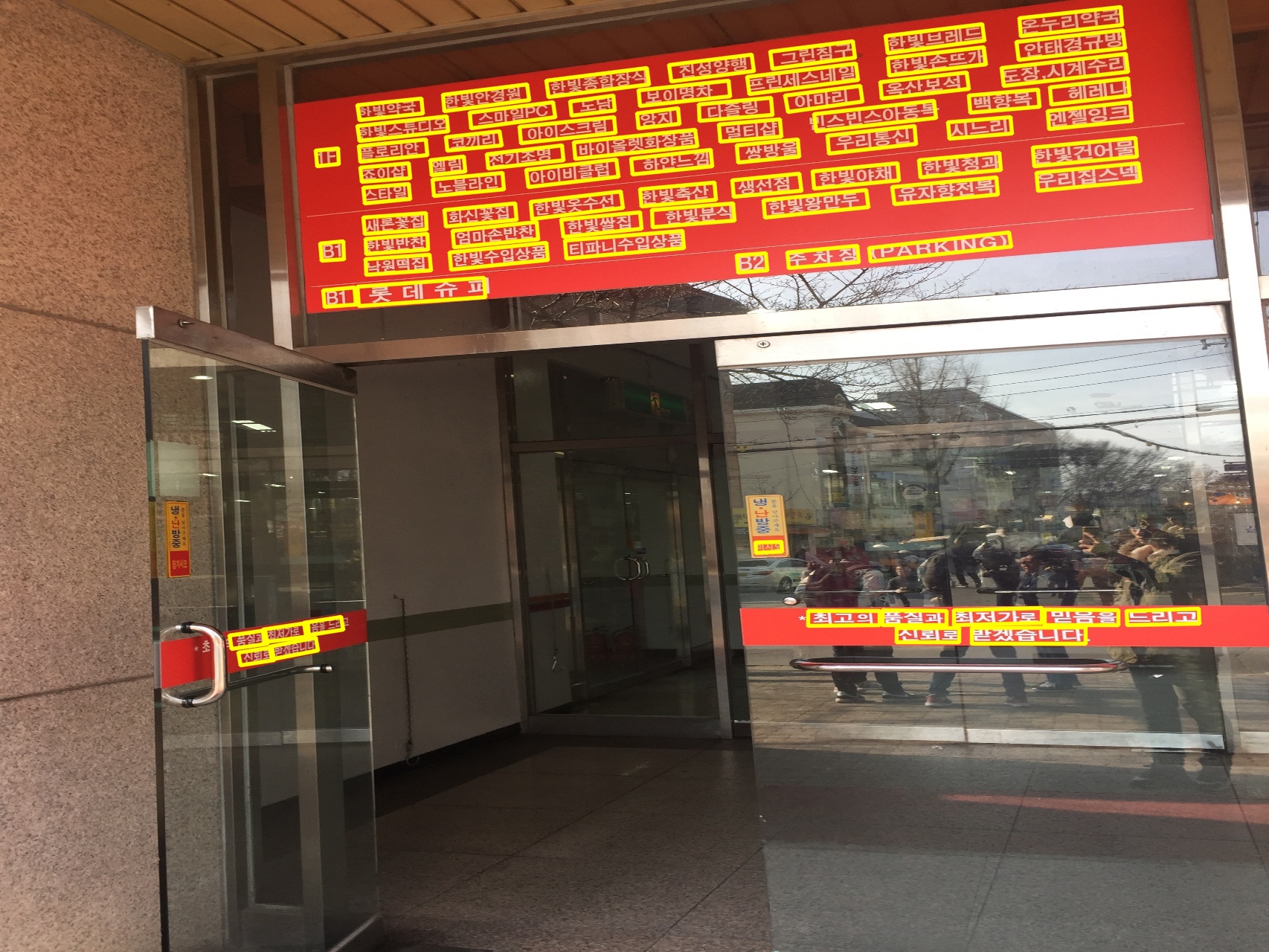} \\
\vspace{-10pt}
\includegraphics[width=\textwidth]{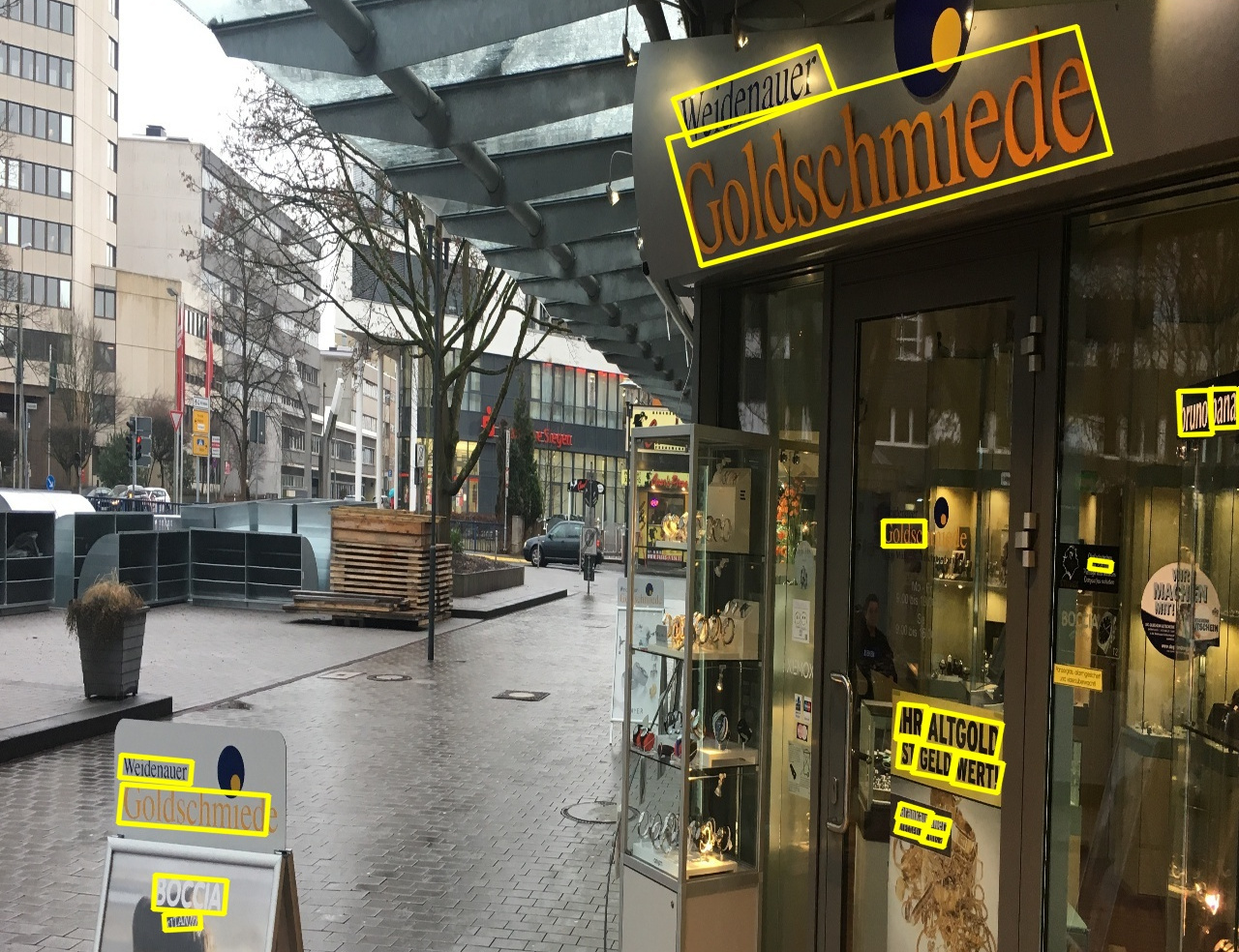}
\end{minipage}
}
\subfigure[ICDAR 2013]{
\begin{minipage}[b]{0.225\textwidth}
\includegraphics[width=\textwidth]{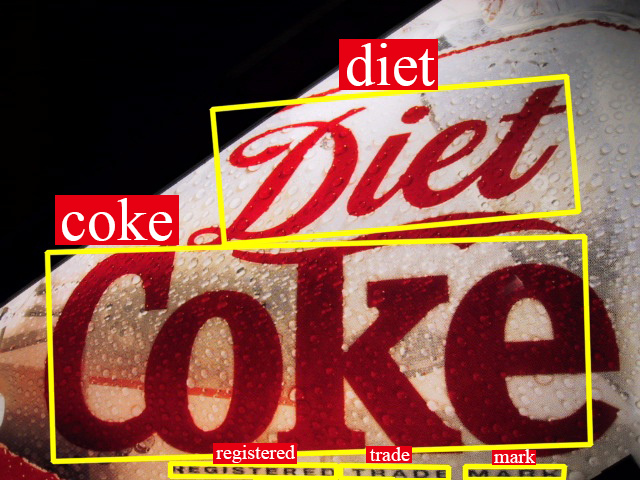} \\
\vspace{-10pt}
\includegraphics[width=\textwidth]{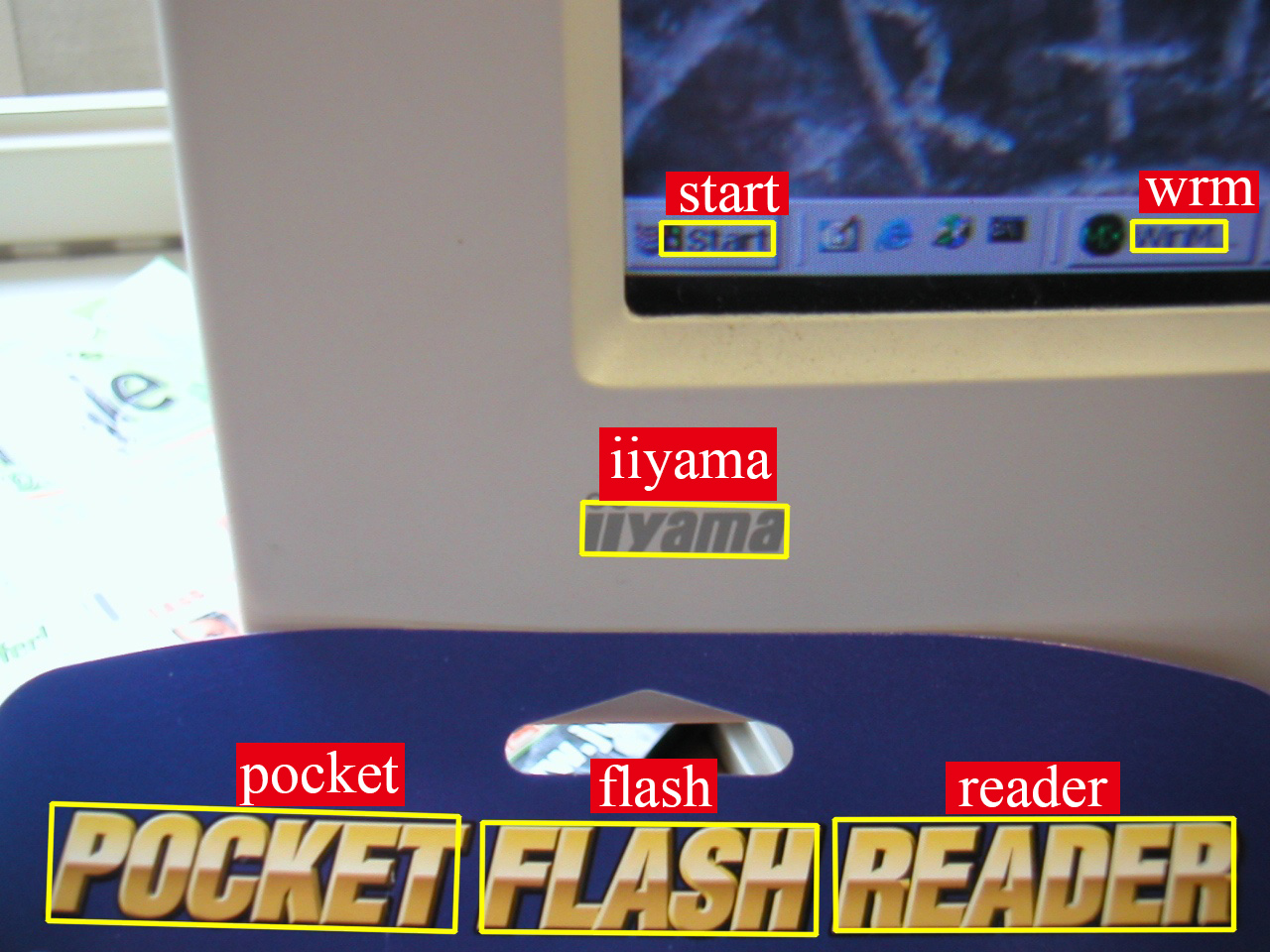}
\end{minipage}
}
\end{center}
\vspace{-10pt}
\caption{Results of the proposed method. Note: we only show text detection results of ICDAR 2017 MLT due to the absence of text spotting task.}
\label{tab:result}
\end{figure*}

FOTS performs better in detection because text recognition supervision helps the network to learn detailed character level features. To analyze in detail, we summarize four common issues for text detection, \textbf{Miss}: missing some text regions, \textbf{False}:  regarding some non-text regions as text regions wrongly, \textbf{Split}: wrongly spliting a whole text region to several individual parts, \textbf{Merge}: wrongly merging several independent text regions together. As shown in Fig. \ref{tab:example}, FOTS greatly reduces all of these four types of errors compared to ``Our Detection" method. Specifically, ``Our Detection" method focuses on the whole text region feature rather than character level feature, so this method does not work well when there is a large variance inside a text region or a text region has similar patterns with its background, etc. As the text recognition supervision forces the model to consider fine details of characters, FOTS learns the semantic information among different characters in one word that have different patterns. It also enhances the difference among characters and background that have similar patterns. As shown in Fig. \ref{tab:example}, for the \textbf{Miss} case, ``Our Detection" method misses the text regions because their color is similar to their background. For the \textbf{False} case, ``Our Detection" method wrongly recognizes a background region as text because it has ``text-like" patterns (\eg, repetitive structured stripes with high contrast), while FOTS avoids this mistake after training with recognition loss which considers details of characters in the proposed region. For the \textbf{Split} case, ``Our Detection" method splits a text region to two because the left and right sides of this text region have different colors, while FOTS predicts this region as a whole because patterns of characters in this text region are continuous and similar. For the \textbf{Merge} case, ``Our Detection" method wrongly merges two neighboring text bounding boxes together because they are too close and have similar patterns, while FOTS utilizes the character level information given by text recognition and captures the space between two words.

\subsection{Comparisons with State-of-the-Art Results}

In this section, we compare FOTS to state-of-the-art methods. As shown in Tab. \ref{tab:icdar15_detect_compare}, \ref{tab:MLT_detect_compare}, \ref{tab:icdar13_detect_compare}, our method outperforms all others by a large margin in all datasets. Since ICDAR 2017 MLT does not have text spotting task, we only report our text detection result. All text regions in ICDAR 2013 are labeled by horizontal bounding box while many of them are slightly tilted. As our model is pre-trained using ICDAR 2017 MLT data, it also can predict orientations of text regions. Our final text spotting results keep predicted orientations for better performance, and due to the limitation of the evaluation protocol, our detection results are the minimum horizontal circumscribed rectangles of network predictions. It is worth mentioning that in ICDAR 2015 text spotting task, our method outperforms previous best method \cite{shi2017seglink,shi2016crnn} by more than 15\% in terms of F-measure.

For single-scale testing, FOTS resizes longer side of input images to 2240, 1280, 920 respectively for ICDAR 2015, ICDAR 2017 MLT and ICDAR 2013 to achieve the best results, and we apply 3-5 scales for multi-scale testing.

\begin{table}
\small
\setlength{\tabcolsep}{3pt}
\begin{center}
\begin{tabular}{c|l|c|c|c}
\hline
\multirow{2}{*}{Dataset} & \multirow{2}{*}{Method} & \multicolumn{2}{c|}{Speed} & \multicolumn{1}{c}{\multirow{2}{*}{Params}} \\ \cline{3-4}
           &             & \multicolumn{1}{c|}{Detection}  & \multicolumn{1}{c|}{End-to-End} & \\ \hline
IC15 & Our Two-Stage      & 7.8 fps         & 3.7 fps         & 63.90 M         \\
   & FOTS     & 7.8 fps         & 7.5 fps         & 34.98 M         \\
   & FOTS RT    & 24.0 fps         & 22.6 fps        & 28.79 M         \\ \hline
IC13 & Our Two-Stage      & 23.9 fps        & 11.2 fps        & 63.90 M         \\
   & FOTS     & 23.9 fps         & 22.0 fps        & 34.98 M         \\ \hline
\end{tabular}
\end{center}
\caption{Speed and model size compared on different methods. ``Our Two-Stage'' consists of a detection model with 28.67M parameters and a recognition model with 35.23M parameters.}
\label{tab:speed_model_size}
\end{table}
\subsection{Speed and Model Size}
\label{speed}
As shown in Tab. \ref{tab:speed_model_size}, benefiting from our convolution sharing strategy, FOTS can detect and recognize text jointly with little computation and storage increment compared to a single text detection network (7.5 fps vs. 7.8 fps, 22.0 fps vs. 23.9 fps), and it is almost twice as fast as ``Our Two-Stage'' method (7.5 fps vs. 3.7 fps, 22.0 fps vs. 11.2 fps). As a consequence, our method achieves state-of-the-art performance while keeping real-time speed.

All of these methods are tested on ICDAR 2015 and ICDAR 2013 test sets. These datasets have 68 text recognition labels, and we evaluate all test images and calculate the average speed. For ICDAR 2015, FOTS uses 2240$\times$1260 size images as inputs, ``Our Two-Stage" method uses 2240$\times$1260 images for detection and 32 pixels height cropped text region patches for recognition. As for ICDAR 2013, we resize longer size of input images to 920 and also use 32 pixels height image patches for recognition. To achieve real-time speed, ``FOTS RT" replaces ResNet-50 with ResNet-34 and uses 1280$\times$720 images as inputs. All results in Tab. \ref{tab:speed_model_size} are tested on a modified version Caffe \cite{jia2014caffe} using a TITAN-Xp GPU.

\section{Conclusion}

In this work, we presented FOTS, an end-to-end trainable framework for oriented scene text spotting. A novel RoIRotate operation is proposed to unify detection and recognition into an end-to-end pipeline. By sharing convolutional features, the text recognition step is nearly cost-free, which enables our system to run at real-time speed. Experiments on standard benchmarks show that our method significantly outperforms previous methods in terms of efficiency and performance.

{\small
\bibliographystyle{ieee}
\bibliography{egbib}
}

\end{document}